%% file: main.tex
\documentclass[letterpaper]{article}
\usepackage{aaai20}
\usepackage{times}
\usepackage{helvet}
\usepackage{courier}
\usepackage[hyphens]{url}
\usepackage{graphicx}
\usepackage{hyperref}       
\usepackage{amsmath}        
\usepackage{graphicx}
\usepackage{epsfig}
\usepackage{subcaption}
\usepackage{tikz}
\usetikzlibrary{bayesnet}

\usepackage{multirow}
\title{Label Dependent Deep Variational Paraphrase Generation}
\author{\Large \textbf{Siamak Shakeri\thanks{corresponding author}, Abhinav Sethy}\\ 
\textsuperscript{}Amazon Alexa AI\\ 
siamaks,sethya@amazon.com
}

\begin{document}
\maketitle
\input{abstract}
\input{intro}
\input{model}
\input{experiments}
\input{results}
\input{conclusion}
%
\clearpage
\bibliographystyle{aaai}
\bibliography{references}

\end{document}

%% file: abstract.tex
\begin{abstract}
Generating paraphrases that are lexically similar but semantically different is a challenging task. Paraphrases of this form can be used to augment data sets for various NLP tasks such as machine reading comprehension and question answering with non-trivial negative examples. In this article, we propose a deep variational model to generate paraphrases conditioned on a label that specifies whether the paraphrases are semantically related or not.  We also present new training recipes and KL regularization techniques that improve the performance of variational paraphrasing models. Our proposed model demonstrates promising results in enhancing the generative power of the model by employing label-dependent generation on paraphrasing datasets.
\end{abstract}
 

%% file: intro.tex
\section{Introduction}\label{sec:intro}
Paraphrase generation refers to the task of generating a sequence of tokens given an input sequence while preserving the overall meaning of the input. Extracting paraphrase from various English translations of the same text is explored in \cite{paraphrase_extract}. Multiple sequence alignments approach is proposed in  \cite{paraphrase_seqalighment} to learn paraphrase generation from unannotated parallel corpora. Deep learning-based paraphrasing has gained momentum recently. Text generation from continuous space using Variational Autoencodes \cite{vae} is proposed in \cite{bowman}. Authors in \cite{paraphraseVAE} suggest using VAEs in paraphrase generation. The ability of VAE generative models at producing diverse sequences makes them a suitable candidate for paraphrasing tasks \cite{creativity}. 

Several publicly available paraphrasing datasets such as \textit{Quora Question Pairs} \cite{quora} and \textit{Microsoft Research Paraphrasing Dataset} \cite{msrp} include a binary label indicating whether the paraphrase sequence is semantically different from the original sentence. Table \ref{tab:dataset_samples} shows samples from \textit{Quora Question Pairs}. Comparing the last two rows of the table, the first row has less common tokens between the original sequence and the paraphrase compared to the second row; however, the paraphrase in the first row conveys the same meaning as the original sequence. This is not the case with the second row, where there is only one token that is different between the paraphrase and the original sequence. We believe that the paraphrase generated when the binary label is 1 follows a different distribution compared to when it is 0. Therefore, including the input label in the neural sequence model would enhance the generative power of such models.

\begin{table*}[h]
\centering
\scriptsize
\begin{tabular}{|c|c|c|}
\hline
Original Sequence&Paraphrase&Label\\ \hline
What is the step by step guide to invest in \textbf{share market in india}? & What is the step by step guide to invest in \textbf{share market}? & 0 \\ \hline
What is the best \textbf{free web hosting for php}? &What are the best \textbf{free web hosting services}?& 0 \\ \hline
How \textbf{will} I open account in Quora? & How \textbf{do} I open an account on Quora? & 1 \\ \hline
\textbf{How should I begin} learning Python? & \textbf{What are some tips for }learning python? &1 \\ \hline
\textbf{What are the possible ways to stop }smoking? & \textbf{How do I quit} smoking? & 1 \\ \hline
What is \textbf{black} hat SEO? & What is \textbf{white} hat SEO? & 0 \\ \hline
\end{tabular}
\caption{Data Samples of Identical versus Non-Identical Paraphrases} 
\label{tab:dataset_samples}
\end{table*}


To the best of our knowledge, paraphrase generation models that take advantage of the label in generating paraphrases have not been explored before. Using variational autoencoders framework to develop models that can produce both semantically \textit{similar} and \textit{dissimiliar} paraphrases is proposed in this article. 


The proposed variational paraphrase generation model is in the family of conditional variational autoencoders of \cite{CVAE}. Further independence assumptions and modifications to the loss function are proposed to the vanilla \textit{CVAE}. Making the generation of hidden variable conditional on the paraphrase label is comparable with  \textit{GMM} priors employed in \textit{TGVAE} model \cite{tgvae}. However, \textit{TGVAE} relies on combined neural topic and sequence modeling in the generative process, while our work assumes the hidden variable being sampled from the \textit{GMM} component that corresponds to the given input label.

The experimental results demonstrate our proposed model outperforms baseline VAE and non-variational sequence to sequence models on the paraphrasing datasets where data samples have a binary label. This label-dependent paraphrase generation can be utilized in extending the size of an already existing training set in various NLP tasks such as question answering, ranking, paraphrase detection.

To summarize, the contributions of this paper are:
\begin{itemize}
    \item We propose label-dependent paraphrase generation for semantically identical and non-identical paraphrasing.
    \item We present a new neural VAE model, DVPG, which benefits from labeled generation as well as variational autoencoding framework.
    \item We suggest several sampling and training schedules that considerably improve the performance of the proposed model.
\end{itemize}
In section \ref{sec:model}, the proposed model is described and its evidence lower bound, also known as \textit{ELBO} \cite{elbo}, is derived. Section \ref{sec:exp} elaborates on  choices in training schedules, variational sampling, model parameters, and measurement metrics. Experimental results are discussed in \ref{sec:results}. Finally, section \ref{sec:conc} summarizes the article and provides future directions for this work.


%% file: model.tex
\section{Model}\label{sec:model}
\begin{figure}[h]
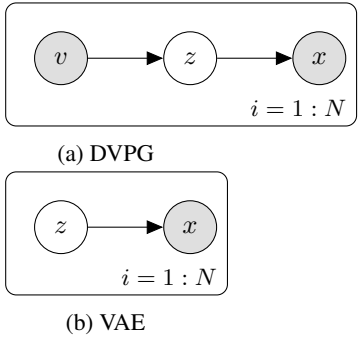

\begin{subfigure}[b]{.15\textwidth}
  \centering
  \tikz{ %
    \node[obs] (v) {$v$} ; %
    \node[latent, right=of v] (z) {$z$} ; %
    \node[obs, right=of z] (x) {$x$} ; 
    \plate[inner sep=0.25cm, xshift=-0.12cm, yshift=0.12cm] {plate1} {(v) (z) (x)}{$i=1:N$}; %
    \edge {v} {z} ; %
    \edge {z} {x} ; %
  }
\caption{DVPG}
\label{figs:graphicalmodel1}
\end{subfigure}%
\\
\begin{subfigure}[b]{.15\textwidth}
  \centering
  \tikz{ %
    \node[latent] (z) {$z$} ; %
    \node[obs, right=of z] (x) {$x$} ; 
    \plate[inner sep=0.25cm, xshift=-0.12cm, yshift=0.12cm] {plate1} {(z) (x)}{$i=1:N$}; %
    \edge {z} {x} ; %
  }
\caption{VAE}
\label{figs:graphicalmodel2}
\end{subfigure}
\caption{DVPG and VAE graphical models}
\label{figs:graphicalmodel}
\end{figure}

The proposed generative model is depicted in Figure \ref{figs:graphicalmodel}. $v$ and $x$ represent observed label and text sequence, respectively. $z$ is the hidden variable. Figures \ref{figs:graphicalmodel1} and \ref{figs:graphicalmodel2} show the proposed model versus vanilla VAE \cite{vae}. We believe the proposed DVPG (\textbf{D}eep Class \textbf{V}ariational \textbf{P}araphrase \textbf{G}eneration) model is more capable than the vanilla VAE in probability density estimation of label-dependent paraphrasing datasets due to the inclusion of label information in the generation of the hidden state.

Generation path of DVPG consists of $ v \sim p(v), z \sim p(z|v), x \sim p(x \vert z)$ and its inference path is as follows: : $v \sim p(v), x \sim p(x), z \sim q(z \vert x, v).$ In the following part, the derivation of the evidence lower bound(ELBO) of the proposed model is explained.

\subsection{Factorization and Objective}
Maximizing the likelihood of the observed variables, $p(x,v)$, is used as the training objective. In the following, derivation and parameterization of the objective function are explained.
\begin{align}
    {\label{eqs:logprob}}
&\log p(x,v) =\sum_{z}q(z \vert x, v)\log p(x,v) \nonumber\\&= \sum_{z}q(z \vert x, v)\log \frac {p(x,v,z)}{p(z\vert x,v)} \nonumber \\ &= \sum_{z}\Big[ q(z \vert x, v) \log \frac{p(x,v,z)}{q(z\vert x, v)} - q(z\vert x, v) \log \frac{p(z\vert x,v)}{q(z \vert x, v)} \Big] \nonumber \\&\geq \sum_{z} q(z \vert x, v) \log \frac{p(x,v,z)}{q(z \vert x, v)}
\end{align}

Where \textit{KL} denotes the Kullback-Leibler divergence. Using the independence assumptions from Figure \ref{figs:graphicalmodel1} :
\begin{align}{\label{eqs:factor}}
p(x,v,z) = p(v)p(z\vert v)p(x \vert z)
\end{align}
Using \ref{eqs:factor}, we can rewrite \ref{eqs:logprob}:
\begin{align*}
&\sum_{z} q(z \vert x, v) \log \frac{p(x,v,z)}{q(z\vert x, v)} =\sum_{z} q(z \vert x, v) \log \frac{p(z \vert v)}{q(z \vert x, v)} + \\&\sum_{z} q(z \vert x, v) \log p(x \vert z) +\sum_{z} q(z \vert x, v) \log p(v) \\
&=-KL(q(z \vert x, v) \vert \vert p(z \vert v)) + E_{z \sim q(z \vert x, v)}\big[p(x \vert z)  + p(v)\big]
\end{align*}
Therefore, the Evidence Lower Bound can be written as :
\begin{align}{\label{eqs:elbo}}
ELBO&=-KL(q(z \vert x, v) \vert \vert p(z \vert v)) + \nonumber \\& E_{z \sim q(z \vert x, v)}\big[p(x \vert z) + p(v)\big]
\end{align}
\subsection{Variational Parameterization} \label{sec:var_params}
In order to simplify the calculation of KL-divergence loss and being able to take advantage of the reparameterization trick \cite{vae}, we made the following assumptions:
\begin{align*}
& q_{\theta}(z \vert x, v) \mapsto \mathcal{N}(\mu^{{\theta}_{\mu,v}},\,\sigma^{{\theta}_{\sigma,v}}) 
,\;\;\; p_{\phi}(z\vert v) \mapsto \mathcal{N}(\mu^{{\phi}_{\mu,v}},\,\sigma^{{\phi}_{\sigma,v}}) 
\end{align*}
The superscripts indicate the parameterization. $\mathcal{N}(\mu,\sigma)$ indicates Gaussian distribution with mean $\mu$, and standard deviation $\sigma$. The entire set of parameters are : $\theta_{\mu,v}, \theta_{\sigma,v}, \phi_{\mu,v}, \phi_{\sigma,v}$, and $\kappa$ where $v \in {0,1}$. 
The optimization problem is to maximize the following: 
\begin{align}{\label{eqs:obj}}
ELBO=-KL(q_{\theta}(z\vert x, v) \vert \vert p_{\phi}(z \vert v)) + E_{z \sim q_{\theta}(z \vert x, v)}\big[p_{\kappa}(x \vert z)\big]
\end{align}
We propose including KL divergence terms to regularize $q_{\theta}$ and $p_{\phi}$ to avoid degeneration of those pdfs.
With regularization terms:
\begin{align}{\label{eqs:obj_reg}}
ELBO&=E_{z \sim q_{\theta}(z \vert x, v)}\big[p_{\kappa}(x \vert z)\big] 
-KL(q_{\theta}(z\vert x, v) \vert \vert p_{\phi}(z \vert v)) \nonumber \\&  
-KL(q_{\theta}(z\vert x, v) \vert \vert \mathcal{N}(0,1)) -KL(p_{\phi}(z \vert v) \vert \vert \mathcal{N}(0,1))
\end{align}
Since $p(v)$ is known, its term is removed from the equation (\ref{eqs:elbo}), and not included in further derivations of ELBO.
During the training of the model, where the objective in \ref{eqs:obj_reg} is maximized, the following path is followed for each $(x,x',v)$ instance:
$z \sim q_{\theta}(z \vert x, v), x'' \sim p_{\kappa}(x \vert z)$. $x''$ and $x'$ are used as the prediction of the model and ground truth, respectively, to compute the cross-entropy loss and other measurement metrics (section \ref{sec:metrics}). Following this approach during the training, $p_{\phi}(z \vert v)$ term does not appear in the training path. Therefore, a modified ELBO can be formed by setting $p_{\phi}(z \vert v) \mapsto \mathcal{N}(0,1)$. This will result in the following:
\setlength\itemsep{0em}
\begin{align}{\label{eqs:obj_reg_2}}
ELBO &= E_{z \sim q_{\theta}(z \vert x, v)}\big[p_{\kappa}(x \vert z)\big] \nonumber \\& -2 \times KL(q_{\theta}(z\vert x, v) \vert \vert \mathcal{N}(0,1))  
\end{align}
Experiments using equations \ref{eqs:obj}, \ref{eqs:obj_reg} and \ref{eqs:obj_reg_2} as ELBO were performed and results reported in sections \ref{sec:exp} and \ref{sec:results}.

\begin{figure*}[]
\centering
\begin{subfigure}{.51\textwidth}
  \centering
  \includegraphics[width=1.1\linewidth]{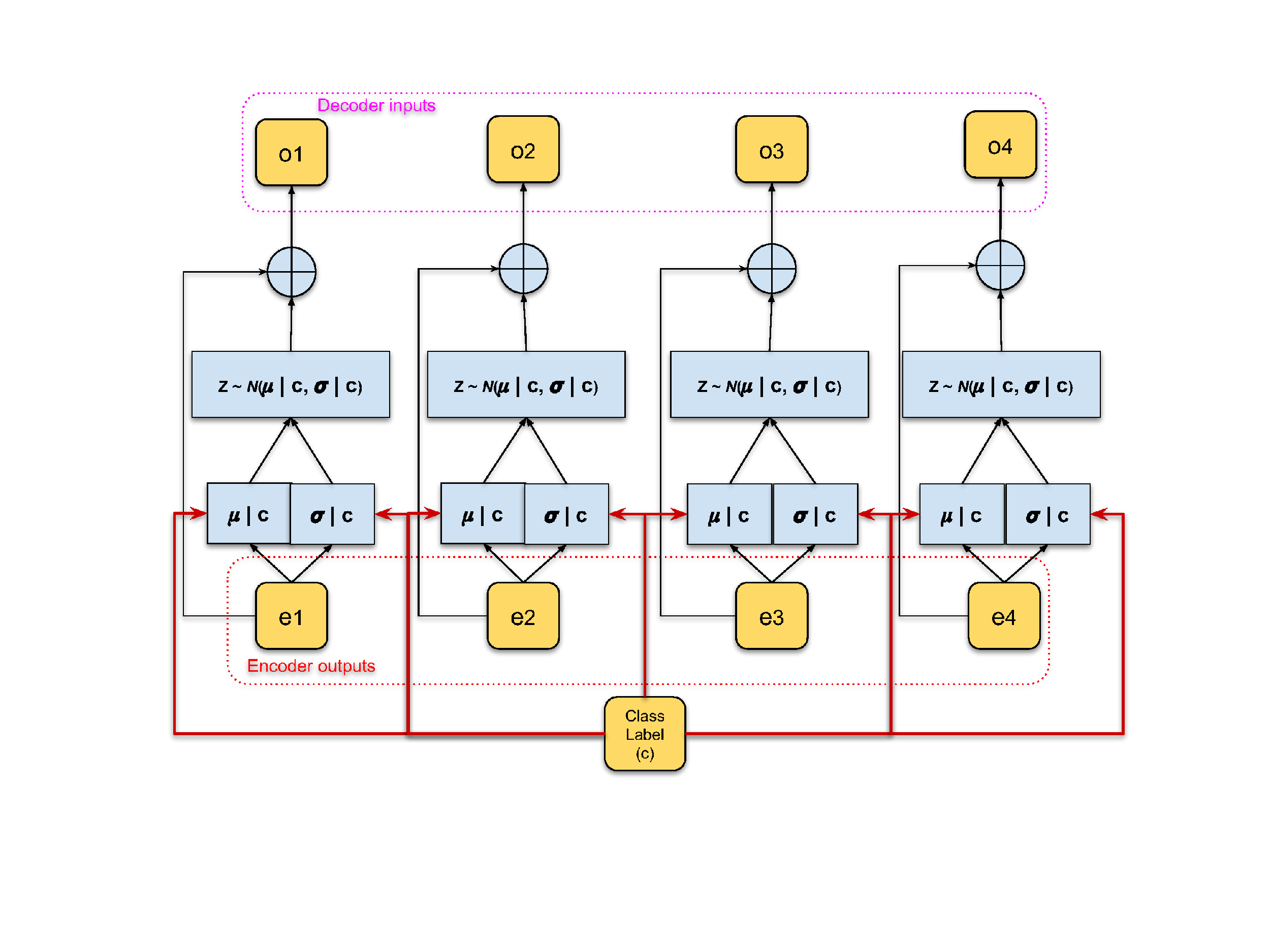}
  \caption{Independent}
  \label{fig:sampling_avg}
\end{subfigure}%
\begin{subfigure}{.5\textwidth}
  \centering
  \includegraphics[width=1.1\linewidth]{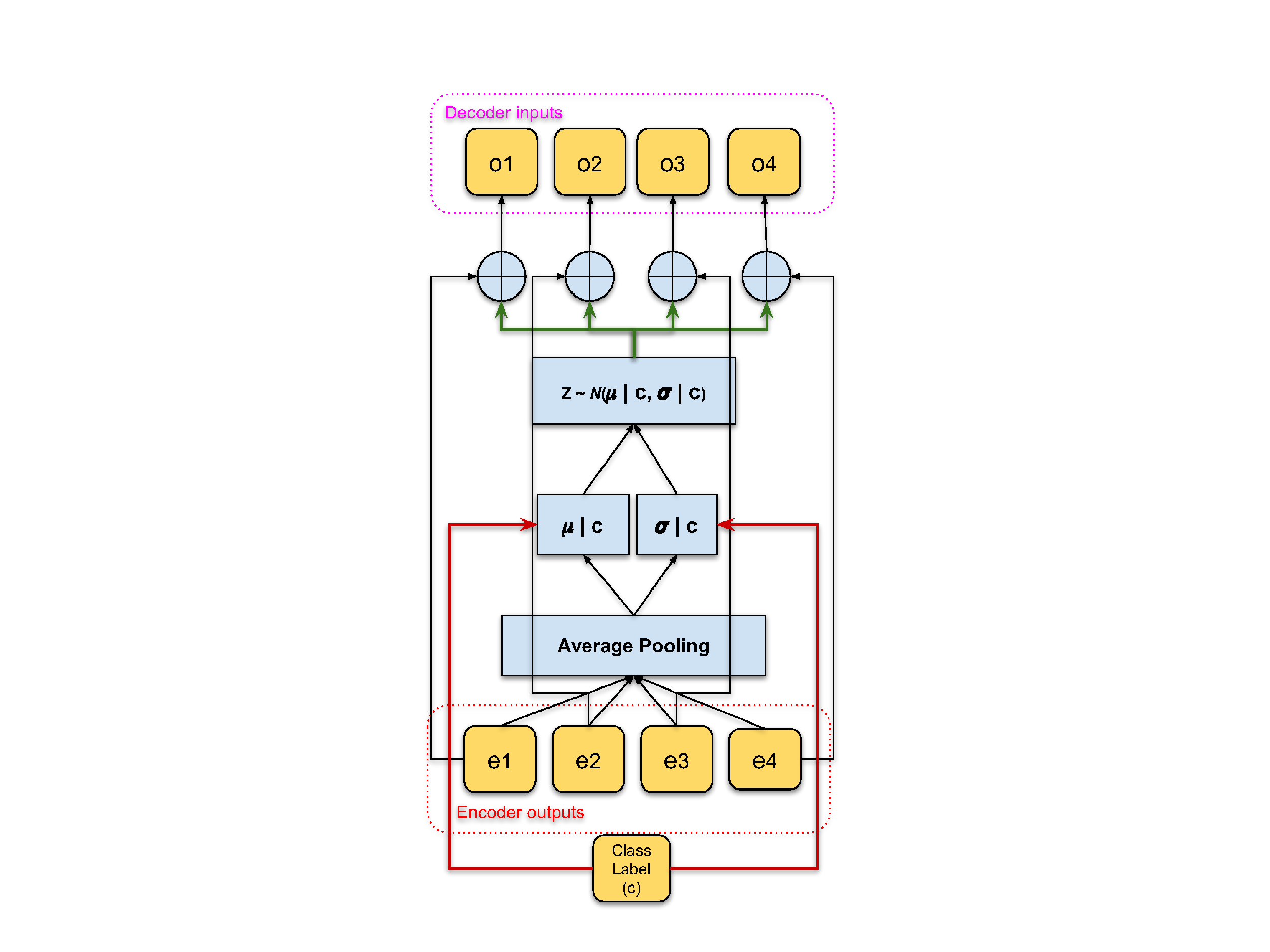}
  \caption{Aggregated}
  \label{fig:sampling_best}
\end{subfigure}
\caption{Independent vs Aggregated Labeled Variational Sampling in DVPG}
\label{fig:dvpg_sampling}
\end{figure*}

%% file: experiments.tex
\section{Experiments}\label{sec:exp}
The baseline neural network used in the experiments is the CopyNet sequence to sequence model introduced by \cite{copynet}. We chose this model due to its ability in selecting sub-phrases of the input sentence to be included in the output \cite{pointer_summerization}. Since paraphrasing requires the generation of a sequence that is lexically similar to the input sequence, the CopyNet model would be a fitting choice \cite{rl_paraphrase}. 

The encoding layer consists of applying Transformer network \cite{transformer} to \textit{BERT}\cite{bert} contextualized word embeddings of the input sequence. An LSTM \cite{lstm} decoder augmented with copy mechanism and cross attention over the encoder outputs performs the generation of the paraphrase sequence.

The following models, loss types and training schedules were explored to measure the performance of the proposed approach:
\subsection{Models} \label{sec:models}
\begin{itemize}
  \setlength\itemsep{0em}
    \item \textit{VAE}: Vanilla VAE as in \cite{vae}.
    \item \textit{Baseline}: Non-variational CopyNet baseline as in \cite{copynet}.
    \item \textit{DVPG}: \textbf{D}eep \textbf{V}ariational \textbf{P}araphrase \textbf{G}eneration model proposed in this work.
\end{itemize}

\subsection{Losses} \label{sec:kld_losses}
The lower bound of likelihood, derived in \ref{eqs:elbo}, consists of a cross-entropy term and KL-divergence term, which will be referred to as \textit{KL}. The proposed variations of the KL term, as derived in \ref{eqs:obj_reg} and \ref{eqs:obj_reg_2}, are enumerated as below:
\begin{itemize}
\item \textit{Loss 1}: KL loss in equation \ref{eqs:obj_reg_2}.
\item \textit{Loss 2}: KL loss in equation \ref{eqs:elbo} without any regularization terms added.
\item \textit{Loss 3}: KL loss in equation  \ref{eqs:obj_reg}.
\end{itemize}

\subsection{Training Schedules} \label{sec:training_schedules}
Avoiding mode collapse is one of the challenges when training variational autoencoders. \cite{vae} suggest KL cost annealing to mitigate this issue, where the KL term is multiplied by a coefficient which gradually increases from zero to one as the training progresses. We employed this method in training all of the variational models in this work.

Curriculum training proposed by \cite{curriculum} is experimented with, where for a fixed number of batches at the beginning of the training, we discard the variational variables and KL loss. Therefore, the model is trained as a non-variational encoder-decoder. After the model is trained with the fixed number of batches, the variational variables and KL term are added. KL coefficient annealing is applied as well. This curriculum learning scheme divides the training into two distinct phases: \textit{CE training}, where the decoder language model, copy mechanism and encoder are well trained to fit the training data, and \textit{variational training}, where the prior is trained. The experimental results show this approach combined by cost annealing outperforms the rest. This training schedule is referred to as \textit{two-step} in this article.

\subsection{Variational Sampling} \label{sec:var_sampling}
Similar to \cite{paraphraseVAE}, the summation of original encoder outputs of the CopyNet and sampled variational variables (\textit{z}) is used in the decoder and copy mechanism to generate the output. Since CopyNet requires the encodings of each of the non-masked input tokens to generate the output tokens, two approaches are proposed to sample \textbf{z}:
\begin{itemize}
    \item \textit{Independent}: a sample is obtained for each of the input token encodings independently, resulting in \textbf{z} $\in R^{d\times H}$. 
    \item \textit{Aggregated}: The encoder outputs are aggregated by average pooling and \textbf{z} is sampled from the resulting aggregated vector(\textbf{z} $\in R^{H}$). 
\end{itemize}

$H$ denotes the hidden dimension and $d$ the length of the non-masked input sequence. Figure \ref{fig:dvpg_sampling} visualizes the two proposed approaches when applied to DVPG. 

\subsection{Data}
Data samples are a set of tuples: $(x,x^{\prime},v)$, where $x$ is the original sequence, $x^{\prime}$ the paraphrase of $x$ and $v \in {0,1}$ is the label indicating whether the paraphrase is semantically identical or not. Quora question pairs dataset \cite{quora} is used. This dataset consists of 400K tuples, where each tuple consists of a pair of questions and a label. Although sanitation methods have been applied to this dataset, the ground truth labels are noisy. Only the pairs where the length of both $x^{\prime}$ and $x$ are less than \textit{14} after being tokenized by WordPiece tokenizer \cite{wordpiece}, are selected. This is done to reduce training time. Besides, since the dataset is noisy, we observed that limiting it to shorter phrases would improve the quality. The resulting training, development, and test sets include 97k, 21k, and 21k pairs, respectively. Those pairs that are labeled 0 are not entirely different questions, but questions where only a small fraction of tokens is different.

\subsection{Training Parameters}
\textbf{AllenNLP} \cite{allennlp} and \textbf{PyTorch} \cite{torch} are used as the development and experimentation environments. \textbf{ADAM} optimizer \cite{adam} with learning rate of $10^{-4}$ is used for training.  Transformer encoder consists of \textit{1} layer and \textit{8} attention heads. Projection, feedforward and, hidden dimensions of the encoder are \textit{256}, \textit{128} and, \textit{128}, respectively. Target vocabulary size is pruned to include only the top \textit{5000} frequent tokens, tokenized by WordPiece. Number of decoding steps is limited to maximum input sequence length of \textit{13} tokens. Target embedding dimension is set to \textit{768}. During evaluation decoding, Beam search of size \textit{16} is used. Each of the models have approximately \textit{7} million parameters. We chose this set of parameters such that the baseline model would perform well on the dataset.
Models are trained for \textit{20} epochs, and the best model is chosen based on \textit{Max-BLEU} score on the development set. 

All the hyperparameters are fixed during the training of all the models, and no parameter or hyperparameter tuning is done. The training is done on Amazon EC2 using \textit{p3.16xlarge} instances, which have Tesla V100 GPUs.
\begin{table*}[h]
\centering
\scriptsize
\begin{tabular}{|c|c|c|c|c|c|c|}
\hline
Method&Model&Max-BLEU&Min-TER&Max-ROUGE-1&Max-ROUGE-2&Max-ROUGE-3\\ \hline
\multirow{4}{*}{Type I}&DVPG Loss 3&37.10$\pm$0.27&45.39$\pm$0.08&61.43$\pm$0.24&41.17$\pm$0.15&28.41$\pm$0.07 \\
&DVPG Loss 2&36.68$\pm$0.28&45.50$\pm$0.17&61.16$\pm$0.26&40.87$\pm$0.21&28.20$\pm$0.24 \\
&DVPG Loss 1&36.98$\pm$0.20&45.46$\pm$0.10&61.36$\pm$0.12&41.06$\pm$0.08&28.27$\pm$0.05 \\ 
&VAE&37.041$\pm$0.17&45.4$\pm$0.09&61.32$\pm$0.09&41.03$\pm$0.13&28.25$\pm$0.19 \\ \hline
\multirow{4}{*}{Type II}& DVPG Loss 3&36.61$\pm$0.34&45.35$\pm$0.31&60.45$\pm$0.14&40.22$\pm$0.13&27.38$\pm$0.11 \\
&DVPG Loss 2&37.82$\pm$0.10&44.42$\pm$0.22&61.39$\pm$0.06&41.31$\pm$0.05&28.42$\pm$0.09\\ 
&DVPG Loss 1&36.88$\pm$0.18&45.3$\pm$0.14&60.83$\pm$0.22&40.63$\pm$0.27&27.77$\pm$0.32 \\ 
&VAE&36.87$\pm$0.48&45.24$\pm$0.32&60.85$\pm$0.26&40.49$\pm$0.31&27.55$\pm$0.22 \\ \hline
\multirow{4}{*}{Type III}&DVPG Loss 3&36.04$\pm$0.08&46.11$\pm$0.09&61.23$\pm$0.07&40.61$\pm$0.15&27.63$\pm$0.21 \\
&DVPG Loss 2&35.72$\pm$0.12&46.17$\pm$0.07&60.67$\pm$0.22&40.27$\pm$0.14&27.56$\pm$0.11 \\ 
&DVPG Loss 1&35.94$\pm$0.06&46.20$\pm$0.13&61.08$\pm$0.16&40.54$\pm$0.03&27.63$\pm$0.14 \\
&VAE&35.85$\pm$0.16&46.27$\pm$0.19&61.22$\pm$0.12&40.58$\pm$0.07&27.61$\pm$0.13\\ \hline
\multirow{4}{*}{Type IV}&DVPG Loss 3&38.13$\pm$0.13&44.46$\pm$0.24&\textbf{62.58}$\pm$0.34&41.97$\pm$0.16&28.75$\pm$0.13\\
&DVPG Loss 2&\textbf{38.42}$\pm$0.19&\textbf{44.09}$\pm$0.26&62.55$\pm$0.43&\textbf{42.10}$\pm$0.27&\textbf{28.92}$\pm$0.23\\
&DVPG Loss 1&38.33$\pm$0.11&44.20$\pm$0.14&62.52$\pm$0.30&42.03$\pm$0.21&28.84$\pm$0.16\\ 
&VAE&38.03$\pm$0.42&44.42$\pm$0.28&62.21$\pm$0.47&41.73$\pm$0.37&28.6$\pm$0.36\\ \hline
\multicolumn{2}{|c|}{Best Model}&DVPG&DVPG&DVPG&DVPG&DVPG \\ \hline
\multicolumn{2}{|c|}{Best Loss Type}&2&2&3&2&2\\ \hline
\multicolumn{2}{|c|}{Best Training Type}&IV&IV&IV&IV&IV \\ \hline
-&Seq2Seq Baseline&29.53$\pm$0.08&51.46$\pm$0.04&56.60$\pm$0.27&35.29$\pm$0.14&22.79$\pm$0.10\\ \hline

\end{tabular}
\caption{Comparison of Best-\textit{metric} scores. Average and standard deviation of results are calculated over three runs of each experiment with different initial seeds.}
\label{tab:best_metrics}
\end{table*}
\subsection{Metrics} \label{sec:metrics}
Metrics frequently used in text generation applications such as machine translation \cite{nmt}, summarization \cite{summarization} and paraphrasing are employed to measure the performance of the models. They are as follows: \textbf{ROUGE-1}, \textbf{ROUGE-2}, \textbf{ROUGE-3}, \textbf{BLEU-4}, and \textbf{TER}.
As suggested in \cite{creativity}, generating only 1 sample for each paraphrase tuple and calculating the metrics, as described above, does not reasonably demonstrate the generative power of variational models. The variational variable \textbf{z}, as discussed in section \ref{sec:var_params}, would encourage the decoder to generate sentences that are token-wise and semantic-wise more diverse when compared to the baseline sequence to sequence model. This could lead to lower performance when compared with the non-variational baseline. 
One approach suggested in \cite{creativity} is to generate multiple paraphrase sequences for each input sequence, and measure the best performing sequence based on the selected criteria, therefore, letting the generative model more chances of generating a paraphrase that matches the oracle sequence more closely.

Following this approach, during the evaluation on development and test sets, for each input tuple, a fixed number of paraphrases is generated, and the following values are calculated for each of the metrics discussed in \ref{sec:metrics}
\begin{itemize}
    \item \textbf{Avg-\textit{metric}}: for each generated sample, the desired \textit{metric} is measured with respect to the reference paraphrase, and the average is calculated over all the generated samples.
    \item \textbf{Best-\textit{metric}}: Similar to Avg-\textit{metric}, except the sequence showing the best performance with respect to the \textit{metric} is selected and used in calculating the desired \textit{metric}. This is referred to in \cite{oracle_metrics} as \textit{Oracle} metric.
\end{itemize}

%% file: results.tex
\section{Results} \label{sec:results}

Experiments were done for the models in section \ref{sec:models}, losses in section \ref{sec:kld_losses}, training schedules in section \ref{sec:training_schedules}, and variational sampling discussed in \ref{sec:var_sampling}.
When performing \textit{two-step} training, only \textit{CE} minimization is performed for the first \textit{6} epochs, after which variational variable and \textit{KL} loss minimization are also included in the training process. We chose this number because we observed that after \textit{6} epochs, the non-variational baseline achieves competitive \textit{BLEU} score on the development set.
During the evaluation of variational models, \textit{10} samples are generated for each development and test set tuple when calculating Max-metrics and Avg-metrics. Each experiment is performed with \textit{3} different seeds. Average and standard deviation of each of the metrics is calculated and compared. 

Configurations used in training the variational models are enumerated as follows: \textbf{Type I}: \textit{Independent} variational sampling, \textbf{Type II}: \textit{Independent} variational sampling + \textit{two-step} training, \textbf{Type III}: \textit{Aggregated} variational sampling, \textbf{Type IV}: \textit{Aggregated} variational sampling + \textit{two-step} training. When reporting performance of \textbf{DVPG} models, the applied KL loss (\ref{sec:kld_losses}) is appended to the model tag.
\begin{figure*}[]
\centering
\begin{subfigure}{.5\textwidth}
  \centering
  \includegraphics[width=1\linewidth]{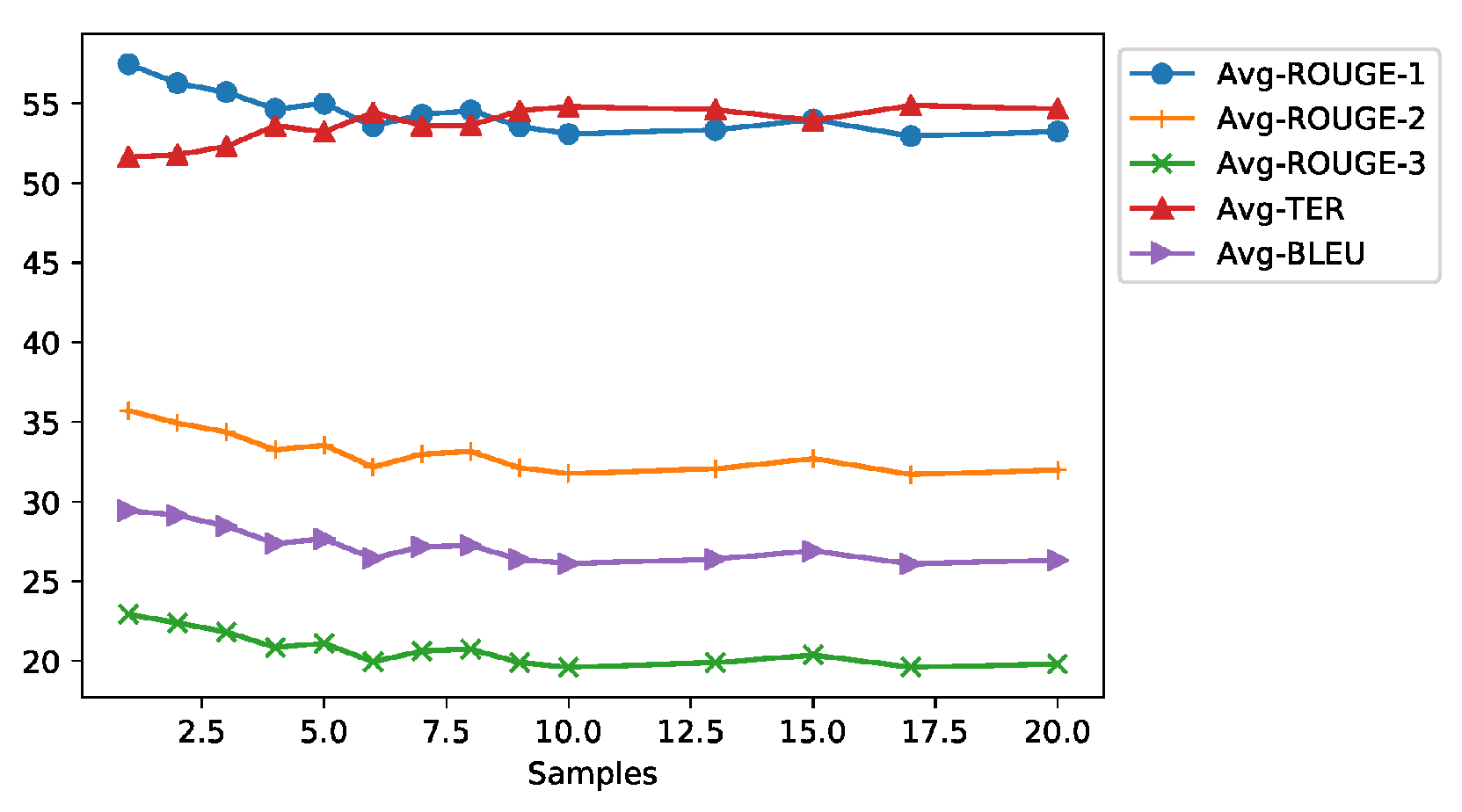}
  \caption{Average Metrics}
  \label{fig:sampling_avg}
\end{subfigure}%
\begin{subfigure}{.5\textwidth}
  \centering
  \includegraphics[width=1\linewidth]{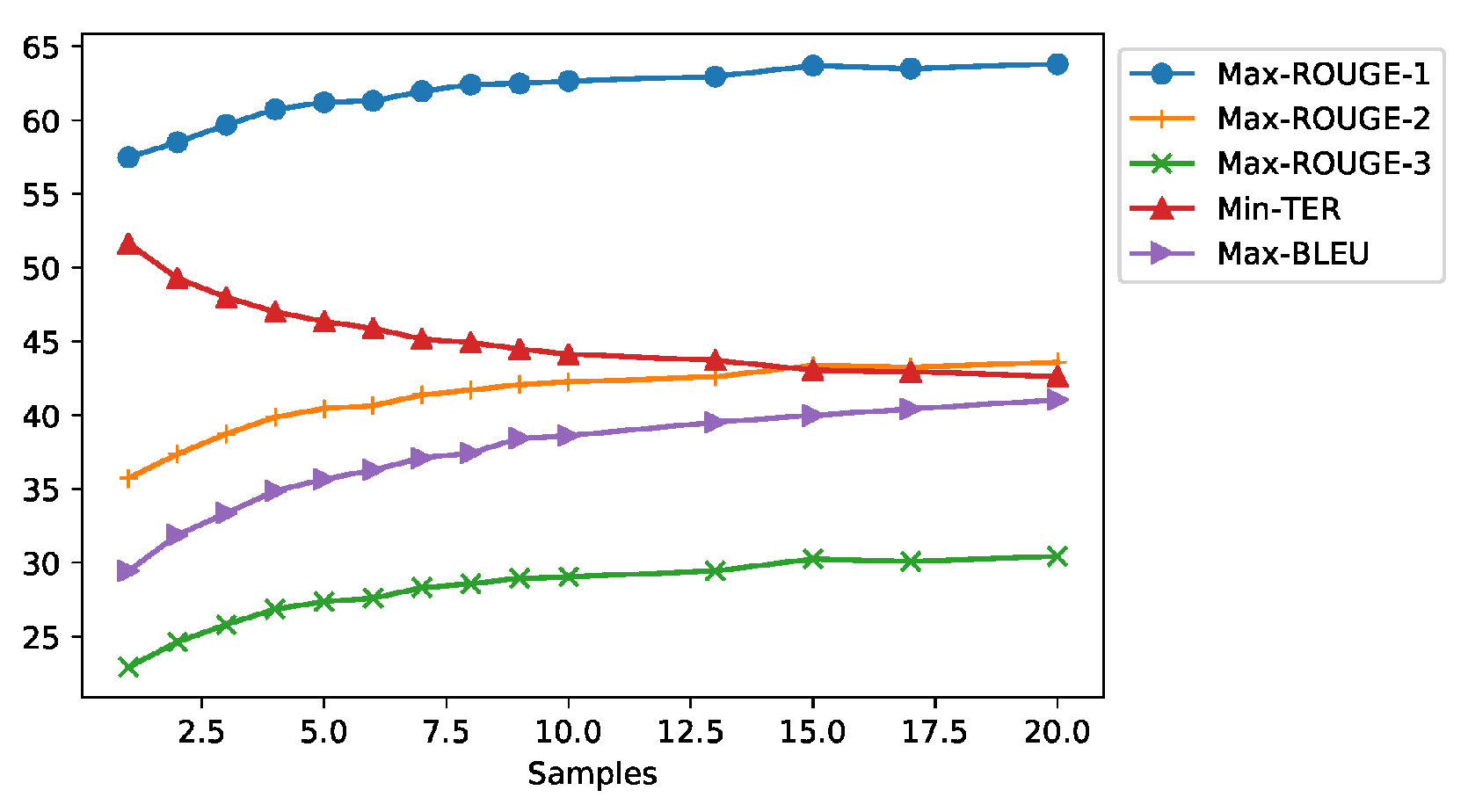}
  \caption{Best Metrics}
  \label{fig:sampling_best}
\end{subfigure}
\caption{Changes to average and best metric values when changing the number of samples DVPG model.}
\label{fig:sampling}
\end{figure*}
The models are trained on the training set; the development set is used to select the best model. \textit{Max-BLEU} is used as the selection metric. Once the best model is selected for each of the settings of loss, training schedule and model type, the model is run over the test set, and the results are reported in Tables \ref{tab:best_metrics} and \ref{tab:avg_metrics}.

\begin{table*}[h]
\scriptsize
\centering
\begin{tabular}{|c|c|c|c|c|c|c|c|}
\hline
Method&Model&Avg-BLEU&Avg-TER&Avg-ROUGE-1&Avg-ROUGE-2&Avg-ROUGE-3&Total Loss\\ \hline
\multirow{4}{*}{Type I}&DVPG Loss 3&28.37$\pm$0.17&52.43$\pm$0.11&55.56$\pm$0.23&34.19$\pm$0.19&21.70$\pm$0.16&16.94$\pm$0.30\\ 
&DVPG Loss 2&28.71$\pm$0.27&52.07$\pm$0.15&55.5$\pm$0.51&34.33$\pm$0.35&21.94$\pm$0.31&\textbf16.53$\pm$0.11\\ 
&DVPG Loss 1&28.49$\pm$0.03&52.33$\pm$0.07&55.63$\pm$0.02&34.31$\pm$0.03&21.82$\pm$0.04&16.99$\pm$0.36\\
&VAE&27.97$\pm$0.29&52.86$\pm$0.38&54.97$\pm$0.20&33.76$\pm$0.20&21.32$\pm$0.24&17.49$\pm$0.89\\ \hline
\multirow{4}{*}{Type II}&DVPG Loss 3&24.10$\pm$1.09&55.74$\pm$0.93&51.01$\pm$1.05&29.48$\pm$1.22&17.49$\pm$1.03&17.93$\pm$0.42\\
&DVPG Loss 2&26.07$\pm$1.03&54.02$\pm$0.85&52.88$\pm$1.17&31.58$\pm$1.11&19.34$\pm$1.02&17.45$\pm$0.33\\
&DVPG Loss 1&25.65$\pm$0.67&54.48$\pm$0.58&52.66$\pm$0.75&31.23$\pm$0.78&19.02$\pm$0.64&17.54$\pm$0.29\\
&VAE &24.53$\pm$0.67&55.79$\pm$0.64&51.37$\pm$0.69&29.96$\pm$0.75&17.96$\pm$0.64&18.0$\pm$0.25\\ \hline
\multirow{4}{*}{Type III}&DVPG Loss 3&28.59$\pm$0.39&52.41$\pm$0.37&55.72$\pm$0.42&34.49$\pm$0.41&21.96$\pm$0.37&18.06$\pm$0.92 \\ 
&DVPG Loss 2&\textbf{29.16}$\pm$0.07&\textbf{51.72}$\pm$0.08&55.8$\pm$0.27&34.76$\pm$0.15&\textbf{22.37}$\pm$0.09&\textbf{16.5}$\pm$0.15\\ 
&DVPG Loss1&28.812$\pm$0.11&52.09$\pm$0.20&55.99$\pm$0.08&34.71$\pm$0.03&22.15$\pm$0.06&17.42$\pm$0.75\\
&VAE&28.82$\pm$0.16&52.15$\pm$0.26&\textbf{56.15}$\pm$0.15&\textbf{34.79}$\pm$0.13&22.19$\pm$0.10&17.6$\pm$0.42\\ \hline
\multirow{4}{*}{Type IV}&DVPG Loss 3&26.37$\pm$0.74&54.82$\pm$0.90&53.38$\pm$1.05&32.14$\pm$0.85&19.92$\pm$0.69&18.88$\pm$0.27\\ 
&DVPG Loss 2&26.02$\pm$0.24&54.95$\pm$0.40&52.86$\pm$0.30&31.66$\pm$0.21&19.52$\pm$0.17&18.71$\pm$0.08\\
&DVPG Loss 1&25.88$\pm$0.27&55.23$\pm$0.43&52.66$\pm$0.33&31.5$\pm$0.26&19.42$\pm$0.20&19.02$\pm$0.04\\
&VAE &25.82$\pm$0.30&55.14$\pm$0.21&52.59$\pm$0.45&31.41$\pm$0.39&19.31$\pm$0.32&19.04$\pm$0.09\\ \hline
\multicolumn{2}{|c|}{Best Model}&DVPG&DVPG&VAE&VAE&DVPG&DVPG \\ \hline
\multicolumn{2}{|c|}{Best Loss Type}&2&2&3&2&2&2 \\ \hline
\multicolumn{2}{|c|}{Best Training Type}&III&III&III&III&III&III \\ \hline
-&Seq2Seq Baseline&29.53$\pm$0.08&51.46$\pm$0.04&56.60$\pm$0.27&35.29$\pm$0.14&22.79$\pm$0.10&15.92$\pm$0.04- \\ \hline
\end{tabular}
\caption{Comparison of Avg-\textit{metric} scores. Average and standard deviation of results are calculated over three runs of each experiment with different initial}
\label{tab:avg_metrics}
\end{table*}

\subsection{\textit{Best}-Metrics} \label{sec:best_metrics_results}
Table \ref{tab:best_metrics} shows the \textit{Best}-metric scores for the proposed training types and models compared with the variational and non-variational baseline. The following can be observed:
\begin{table*}[]
\scriptsize
\centering
\begin{tabular}{|c|c|c|c|c|c|c|}
\hline
Model&Max-BLEU&Min-TER&Max-ROUGE-1&Max-ROUGE-2&Max-ROUGE-3\\ \hline
DVPG Loss 3&\textbf{48.24}$\pm$0.97&47.15$\pm$0.4&69.35$\pm$0.20&51.45$\pm$0.42&40.99$\pm$0.66\\ \hline
DVPG Loss 2&46.32$\pm$1.48&46.69$\pm$0.26&69.3$\pm$0.12&51.54$\pm$0.29&41.29$\pm$0.56\\ \hline
DVPG Loss 1&47.08$\pm$1.42&\textbf{46.48}$\pm$0.21&69.32$\pm$0.12&51.67$\pm$0.35&\textbf{41.57$\pm$0.72}\\ \hline
VAE &47.29$\pm$0.85&46.70$\pm$0.49&69.35$\pm$0.19&51.66$\pm$0.39&41.47$\pm$0.9\\ \hline

Seq2Seq Baseline&45.11$\pm$0.67&49.72$\pm$0.62&65.56$\pm$0.95&48.71$\pm$0.79&39.37$\pm$0.59\\ \hline
\end{tabular}
\caption{Comparison of Best-\textit{metric} scores With Microsoft Research Paraphrasing Corpus}
\label{tab:msrp}
\end{table*}

\begin{itemize}
    \item Variational models overperform the non-variational model by a wide margin. Absolute improvements of \textbf{9\%} in \textit{BLEU}, \textbf{7.4\%} in \textit{TER}, \textbf{6\%} in \textit{ROUGE-1}, \textbf{7\%} in \textit{ROUGE-2}, and \textbf{6\%} in \textit{ROUGE-3} are observed. As similarly reported by \cite{creativity}, this indicates the generative power of variational models in producing diverse outputs.
    \item \textbf{DVPG} model overperforms the baseline \textbf{VAE} with decent margins. Absolute improvements of \textbf{0.39\%} in \textit{BLEU}, \textbf{0.33\%} in \textit{TER}, \textbf{0.37\%} in \textit{ROUGE-1}, \textbf{0.37\%} in \textit{ROUGE-2} and \textbf{0.32\%} in \textit{ROUGE-3} are seen when comparing the best \textbf{DVPG} model to the best VAE model. Considering no parameter tuning is done, and the results are averaged, it demonstrates the efficacy of the proposed model.
    \item \textbf{Two-step} training, as discussed in section \ref{sec:training_schedules}, contributes to improvement in Best-metrics when used with aggregate variational samples (\textit{Type IV}), However, when used with independent variational sampling (\textit{Type II}), does not demonstrate consistent gains.
    \item \textbf{Aggregate} variational sampling results generally overperform \textbf{Independent} sampling method, as shown in Table \ref{tab:best_metrics} by comparing training (\textit{Type III, Type IV}) versus (\textit{Type I, Type II}). This supports the hypothesis that the independent assumption underlying \textbf{Independent} variational sampling is not correct with the sequential input, where there are dependencies between the tokens.
\end{itemize}
\subsection{\textit{Average}-metrics}\label{sec:avg_metrics_discuss}
Average-metric values on the test set are shown in Table \ref{tab:avg_metrics}. It is important to note that the best models are not picked based on the best average value. The best model is picked based on the highest \textit{Max-BLEU} score; therefore, they would not necessarily deliver a fair judgment on the superiority of a model versus the other. Besides, the model with larger diversity in generating outputs has higher chances of producing paraphrases that are on average not similar to the ground truth compared to a model which introduces less diversity in generated sequence. Hence, such a diversity-powerful model, while having higher Max-metrics, might suffer from lower Average-metrics. Nonetheless, \textit{Average-metrics} provide a measure of the average quality of the generated paraphrases.

Similar to \textit{Best}-metrics results, \textbf{DVPG} model with Loss 2 performs the best amongst the DVPG-based models in \textbf{Average-metrics}. Additionally, its performances exceeds VAE's in \textit{BLEU} by \textbf{0.34\%}, in \textit{TER} by \textbf{0.43\%}, in \textit{ROUGE-3} by \textbf{0.18\%} and in \textit{Total Loss} by \textbf{1.1}. The \textit{CE Loss} is not normalized by the length of the sequence, thus the large values. Experiments were done with normalizing \textit{CE} by the sequence length, and the results did not demonstrate higher \textit{Max} or \textit{Average metrics}.

\textbf{VAE} performs better in \textit{ROUGE-1} and \textit{ROUGE-2} by \textbf{0.16\%} and \textbf{0.03\%} absolute values, respectively, while the latter is well within the confidence interval. 

Comparing the \textit{Average-metrics} of \textbf{Variational} models against the \textbf{Seq2Seq} Baseline, it can be observed that the non-variational model exceeds the performance of the best variational model. Furthermore, the setting that demonstrated the best performance in \textit{Best-metrics}(DVPG Loss 2 Type IV), is not the same setting that produces the best metrics amongst the variational models(DVPG Loss 2 Type III). This observation is contributed to the trade-off between diversity and average performance, as discussed previously.
\subsection{Generative Power}
To measure the limit of the generative power of proposed variational models, the number of samples used in variational sampling (section \ref{sec:var_sampling}) is changed from 1 to 20 during the evaluation, and the best model for each sample is selected. The change in \textit{Average} and \textit{Best} metrics are depicted in Figures \ref{fig:sampling_best} and \ref{fig:sampling_avg}.  \textbf{DVPG Loss 2 Type IV} (section \ref{sec:models}) is used. Improvement in the \textit{Best}-metrics by increasing the number of samples diminishes for values larger than \textbf{10}. For example, increasing the sample size from \textit{1} to \textit{10}, results in \textbf{9} points increase in \textit{Max-BLEU}, while increasing it from \textbf{10} to \textbf{20} yields \textbf{1.4} point improvement. This indicates the model has reached its generative capacity with respect to the given dataset, and further enhancement of diversity requires changes in the underlying model architecture. A similar trend can be observed in the other \textit{Best}-metrics. 

Looking at the change in \textit{Average}-metrics in Figure \ref{fig:sampling_avg} reinforces argument regarding the trade-off between diversity and average performance. However, the degradation of \textit{Average}-metrics by increasing the sample size is not proportional to the increase in the \textit{Best} metrics. Looking at \textit{TER} as an example, the effect of changing the sample size from 1 to 20 is \textbf{3} absolute points increase in \textit{Average}-TER, compared to the \textbf{10} points decrease in \textit{Best}-TER. Therefore, we can infer from this observation that the generated output sequences, while being diverse, are still close to the gold output sequence. 

\subsection{Microsoft Research Paraphrasing Dataset}
Table \ref{tab:msrp} explains the results of running \textit{Baseline Seq2Seq}, \textbf{DVPG}, and \textbf{VAE} models when using \textbf{Type IV} training on Microsoft Research Paraphrasing Dataset \cite{msrp}. This dataset set is comprised of \textit{5800} tuples, where each tuple consists of an original string, the paraphrased sequence, and a binary label indicating whether the paraphrased sequence is semantically identical to the original sequence. The dataset is split into three sets of \textit{4100}, \textit{850}, \textit{850} tuples as training, development and test sets, respectively. In the interest of being succinct, only \textit{Best}-metrics are shown. 
As it can be observed, the \textbf{DVPG} model overperforms \textbf{VAE} in all of the metrics. Most notably, absolute improvements of \textbf{0.95\%} in \textit{Max}-BLEU, and \textbf{0.22$\%$} in \textit{Min}-TER are achieved by using the proposed generative process. The improvements in \textit{ROUGE} metrics are marginal. We suspect the relatively small size of the dataset diminishes the improved generative power of the \textbf{DVPG} model. Evidence of this conjecture is the smaller gap between variational and non-variational baseline when compared with the results in Tables \ref{tab:best_metrics} and \ref{tab:avg_metrics}, where the much larger \textit{Quora} dataset was employed. 

An interesting observation is that contrary to results with \textit{Quora} dataset, \textbf{DVPG} with more regularized \textbf{KL} losses overperform the un-regularized \textbf{Loss 2}. This can be contributed to the smaller size of the dataset, which makes such regularization more necessary. 

\subsection{Analysis of KL Losses}
When \textit{two-step} schedule is applied, the \textit{CE} only training is done for the first \textit{12000} batches, or \textit{6} epochs, after which the \textbf{KL} loss(es) are included in the loss function. We speculate that minimizing only the \textbf{CE} loss for several epochs, not only facilitates encoding a larger volume of information in the variational parameter but also enhances the decoder. When training complex sequence to sequence models, where training such encoders and decoders would require multiple epochs, \textit{two-step} training would be more effective versus vanilla KL cost annealing \cite{bowman}.


%% file: conclusion.tex
\section{Conclusion}\label{sec:conc}
Paraphrase generation when there are two classes of paraphrases is explored in this article. A new graphical model is introduced, and the corresponding ELBO is derived. Our experiments on Quora Question Pairs and Microsoft Research Paraphrasing Dataset showed that the proposed model outperforms vanilla VAE and non-variational baseline across the metrics that measure the generative power of the models, therefore supporting the hypothesis that label-dependent paraphrase generation can better learn the distribution of the labeled paraphrasing datasets. Furthermore, the proposed variational sampling and training schedules showed consistent improvements with the variational models.
 One future direction of this work is to explore the setting where the label variable $v$ is not observed, therefore extending its application to unannotated paraphrases corpora. Applying it to NLP tasks such as machine reading comprehension or answer ranking is another continuation of this work. 